\documentclass[letterpaper, 10 pt, conference]{ieeeconf}
\IEEEoverridecommandlockouts
\usepackage{cite}
\usepackage{amsmath,amssymb,amsfonts}
\usepackage{algorithmic}
\usepackage{graphicx}
\usepackage{textcomp}
\usepackage{xcolor}
\usepackage{url}
\usepackage{booktabs}
\usepackage{hyperref}

\hypersetup{
  colorlinks=true,
  linkcolor=blue,
  citecolor=blue,
  urlcolor=blue,
  bookmarksnumbered=true,
  bookmarksopen=true
}

\def\BibTeX{{\rm B\kern-.05em{\sc i\kern-.025em b}\kern-.08em
    T\kern-.1667em\lower.7ex\hbox{E}\kern-.125emX}}

\begin{document}

\title{Benchmarking the Effects of Object Pose Estimation and Reconstruction on Robotic Grasping Success
}

\author{
Varun Burde\textsuperscript{1,2}, Pavel Burget\textsuperscript{1,2}, Torsten Sattler\textsuperscript{1,2}\\[2pt]
\textsuperscript{1}\textit{Faculty of Electrical Engineering, Czech Technical University in Prague, Czechia}\\
\textsuperscript{2}\textit{Czech Institute of Informatics, Robotics and Cybernetics, Czech Technical University in Prague, Czechia}\\
}

\maketitle

\begin{abstract}

3D reconstruction serves as the foundational layer for numerous robotic perception tasks, including 6D object pose estimation and grasp pose generation. Modern 3D reconstruction methods for objects can produce visually and geometrically impressive meshes from multi-view images, yet standard geometric evaluations do not reflect how reconstruction quality influences downstream tasks such as robotic manipulation performance. This paper addresses this gap by introducing a large-scale, physics-based benchmark that evaluates 6D pose estimators and 3D mesh models based on their functional efficacy in grasping. We analyze the impact of model fidelity by generating grasps on various reconstructed 3D meshes and executing them on the ground-truth model, simulating how grasp poses generated with an imperfect model affect interaction with the real object. This assesses the combined impact of pose error, grasp robustness, and geometric inaccuracies from 3D reconstruction. Our results show that reconstruction artifacts significantly decrease the number of grasp pose candidates but have a negligible effect on grasping performance given an accurately estimated pose. Our results also reveal that the relationship between grasp success and pose error is dominated by spatial error, and even a simple translation error provides insight into the success of the grasping pose of symmetric objects. This work provides insight into how perception systems relate to object manipulation using robots.

\end{abstract}

%%%%%%%%%%%%%%%%%%%%%%%%%%%%%%%%%%%%%%%%%%%
\section{Introduction}

The ambition for robots to autonomously operate in human-centric environments is a primary driver of robotics research. A prerequisite for meaningful interaction is the ability to perceive and manipulate objects, which requires both knowing an object's 6D pose (position and orientation) and understanding its geometry. This 3D model serves a basis for model-based 6D pose estimation methods that determine an object's position and orientation and it is the representation upon which grasp poses are generated for physical interaction.
While deep learning has led to remarkable progress in 6D pose estimation \cite{megapose2022, foundationpose2023} and 3D reconstruction \cite{mildenhall2020nerf, wang2021neus}, these perception components are typically evaluated in isolation.

Progress in pose estimation is measured by geometric metrics like ADD (Average Distance of Model Points - Symmetric) on benchmarks like BOP \cite{hodan2018bop}, while reconstruction quality is assessed by metrics such as Chamfer distance. However, this decoupled evaluation creates a significant gap, and it is unclear how errors from pose estimation and geometric reconstruction compound and propagate to affect the success of downstream manipulation tasks like grasping. For a robot, the utility of a perception system is not defined by its geometric precision alone, but by its functional efficacy, an idea shared by other recent benchmarking efforts \cite{graspnet}.

This paper engages this disconnect head-on by evaluating perception systems based on the robot's ability to grasp the object. We introduce a large-scale, systematic study in the physics simulator that directly connects errors from pose estimation and object reconstruction to robotic grasping success. Our key contribution is the analysis of the relation between the manipulation and the reconstruction quality. We propose a new way to evaluate 3D reconstruction methods in the context of object manipulation.

% This paper confronts this disconnect head-on by evaluating perception systems based on their ability to support a functional task. We introduce a large-scale, systematic study in the physics simulator that directly connects errors from both pose estimation and object reconstruction to robotic grasping success. Our key contribution is the analysis of mismatched geometries, where a robot might plan a grasp using an idealized internal model (e.g., a CAD model) but must execute it on the real-world object, which is imperfectly represented by a reconstructed mesh, or vice-versa.

By simulating millions of grasp attempts under these mismatched conditions, we measure the probability of task success as a function of both the underlying pose error and the geometric fidelity of the object model. 
% This unprecedented scale of evaluation reveals critical insights into the practical utility of different perception pipelines that are invisible to traditional, task-agnostic metrics.
 This large-scale evaluation uncovers the hidden flaw of perception pipelines, showing how errors considered negligible by standard metrics can decisively impact downstream grasp execution

% Our contributions are threefold:
% \begin{itemize}
%     \item We introduce a comprehensive framework for functionally evaluating the combined impact of 6D pose estimation and 3D reconstruction errors on robotic grasping.
%     \item We conduct the first large-scale quantitative analysis of grasp success utilizing 3D reconstructed object model for pose estimation and grasp pose generation, revealing the performance degradation caused by geometric inaccuracies.
%     \item We present a task-based re-evaluation of modern perception systems of 3D reconstruction, object pose estimation and Grippiing pose geneartion, providing crucial insights into their practical utility and failure modes for real-world manipulation.
% \end{itemize}

Our contributions are threefold:
\begin{itemize}
    \item We introduce a comprehensive framework for functionally evaluating the combined impact of 6D pose estimation and 3D reconstruction errors on robotic grasping.
    \item We conduct the first large-scale quantitative analysis of grasp success utilizing 3D reconstructed object models for pose estimation and grasp pose generation, revealing the performance degradation caused by geometric inaccuracies.
    \item We present a task-based re-evaluation of modern perception systems, including 3D reconstruction, object pose estimation, and grasp pose generation, providing crucial insights into their practical utility and failure modes for real-world manipulation.
\end{itemize}

% By grounding the evaluation of perception in the physical reality of manipulation, this work seeks to guide future research towards not just geometrically accurate, but functionally robust, perception systems for robotics.

%%%%%%%%%%%%%%%%%%%%%%%%%%%%%%%%%%%%%%%%%%%

\section{Related Work}
Our research is situated at the intersection of 6D object pose estimation, 3D reconstruction, and robotic grasping. We address the critical gap in their unified evaluation by examining how errors from both perception domains propagate to a functional manipulation task.

\subsection{6D Object Pose Estimation and Benchmarks}
Modern 6D pose estimation has shifted from classic feature-based methods \cite{hinterstoisser2012model} to sophisticated learning-based approaches. Methods like PoseCNN \cite{xiang2018posecnn} and DenseFusion \cite{wang2019densefusion} demonstrated the power of deep learning, while recent zero-shot systems like MegaPose \cite{megapose2022} and FoundationPose \cite{foundationpose2023} have achieved remarkable generalization to novel objects. This progress has been accelerated by standardized benchmarks, most notably the BOP challenge \cite{hodan2018bop}, which evaluates methods using task-agnostic geometric metrics like ADD and MSSD (Maximum Symmetry-Aware Surface Distance). While instrumental, these metrics do not capture how pose errors affect physical interaction.

\subsection{3D Reconstruction for Robotics}
Simultaneously, 3D reconstruction from multiview RGB images has seen significant advances, particularly with neural implicit representations. Methods like NeRF (Neural Radiance Fields) \cite{mildenhall2020nerf}, \cite{instantngp} and its variant for implicit representation, such as NeuS \cite{wang2021neus}, Volsdf \cite{volsdf}, Monosdf \cite{bakedsdf} can produce high-fidelity meshes from multiview images. For robotics, these reconstructions serve as the geometric basis for tasks like object pose estimation and grasping. Benchmarks such as \cite{ETH3d, tankandtemples} evaluate the 3D reconstruction in terms of geometric accuracy, and the
\cite{burde2024benchmark} evaluates these methods one step further by evaluating the performance for object pose estimation. However, like pose estimation benchmarks, they do not assess the functional suitability of the resulting meshes for manipulation. A reconstructed mesh with low geometric error might still possess artifacts such as smoothed edges or filled holes that are critical for stable grasping.

\subsection{Robotic Grasping and the Perception-Action Gap}
Robotic grasping research has evolved from analytical, model-based approaches \cite{bicchi2000robotic} to data-driven techniques that learn grasping policies from large datasets \cite{pinto2016supersizing}. While systems like Dex-Net 2.0 \cite{mahler2017dex} show impressive performance by learning robust grasp policies from millions of synthetic examples, they often assume access to high-quality point clouds or object models. The development of common object sets and protocols, such as the widely-used YCB Object and Model Set \cite{YCBV}, provides the necessary foundation for researchers to evaluate the connection between perception and manipulation.

% However, a gap persists in systematically characterizing how the specific errors from modern perception systems (both pose and geometry) affect manipulation. Our work addresses this by directly measuring this relationship, providing a foundational, quantitative analysis of how the quality of the perceived world model impacts functional task success. Instead of proposing new methods to mitigate errors, we provide a rigorous methodology to measure, understand, and characterize the problem's structure, offering an empirical basis for designing the next generation of robust manipulation systems.

However, a gap persists in systematically characterizing how modern perception systems' specific errors (pose and geometry) resulting from a 3D reconstructed model affect manipulation. Our work addresses this by directly measuring this relationship, providing a foundational, quantitative analysis of how the quality of the perceived world model impacts grasping success. Instead of proposing new methods to mitigate errors, we provide a rigorous methodology to measure, understand, and characterize the problem's structure, offering an empirical basis for designing the next generation of robust manipulation systems.

%%%%%%%%%%%%%%%%%%%%%%%%%%%%%%%%%%%%%%%%%%%

\section{Methodology}\label{sec:methodology}

\begin{figure*}[htbp]
  \centering
  \includegraphics[width=\textwidth]{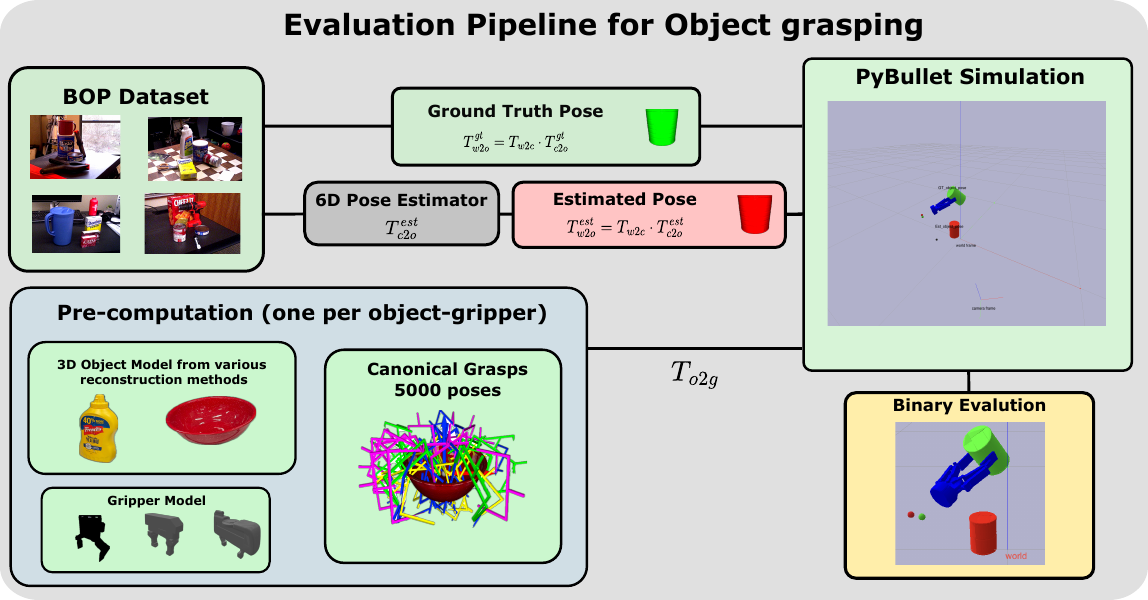}
  % \caption{Overview of the evaluation pipeline. For each object, we first pre-compute a baseline grasp success rate ($S_{gt}$) for nine different grippers. Then, for a given scene, a pose estimate ($T_{c2o}^{est}$) is used to derive a target gripper pose ($T_{w2g}^{est}$). This grasp is executed in a physics simulator on the ground-truth object, and the outcome (success/failure) is recorded and correlated with the initial pose error.}
\caption{Overview of our evaluation pipeline. First, a canonical grasp library is pre-computed for each object. Then, for a given scene, a pose estimator provides $T_{c2o}^{est}$. This pose is used to derive a target gripper pose, $T_{w2g}^{est}$, which is executed on the ground-truth object. The outcome is recorded to calculate the Estimated Success Rate ($S_{est}$) (Sec.~\ref{sec:metric_sest}) and correlated with the initial pose error.}
  \label{fig:pipeline}
\end{figure*}

To systematically quantify how mesh geometric and 6D pose estimation errors propagate to robotic grasping outcomes, we designed a comprehensive benchmarking framework within the PyBullet \cite{coumans2016pybullet} physics simulator. Our methodology is centered around a core transformation chain that links perception to action. We first establish a baseline of ideal grasping performance for a library of grippers and objects. Then, we evaluate the degradation of this performance on a  binary grasp success task. The overall pipeline is visualized in Fig.~\ref{fig:pipeline}.

\subsection{Core Transformation Chain}
The link between perception and robotic action is defined by a sequence of rigid body transformations. Let the primary coordinate frames be World ($W$), Camera ($C$), Object ($O$), and Gripper ($G$). We define the following homogeneous transformations:
\begin{itemize}
    \item $T_{w2c}$: The ground-truth pose of the camera in the world frame, known from the dataset.
    \item $T_{c2o}^{gt}$: The ground-truth pose of the object in the camera frame, provided by the BOP dataset annotations.
    \item $T_{c2o}^{est}$: The pose of the object in the camera frame as predicted by a 6D pose estimation method.
    \item $T_{o2g}$: A pre-computed, canonical grasp pose, defining the gripper's pose relative to the object's local coordinate frame.
\end{itemize}

Using this chain, we can compute the target pose for the gripper in the world frame. The ideal gripper pose, derived from the ground-truth object pose, is:
\begin{equation}
T_{w2g}^{gt} = T_{w2c} \cdot T_{c2o}^{gt} \cdot T_{o2g}
\label{eq:gt_grasp}
\end{equation}
Conversely, the gripper pose that a robot would actually target in a real-world scenario, based on the perception system's output, is:
\begin{equation}
T_{w2g}^{est} = T_{w2c} \cdot T_{c2o}^{est} \cdot T_{o2g}
\label{eq:est_grasp}
\end{equation}
The core of our methodology is to execute grasps using the target pose $T_{w2g}^{est}$ but to evaluate the physical interaction with the object located at its true pose, dictated by $T_{w2c} \cdot T_{c2o}^{gt}$. This setup precisely simulates the real-world scenario where a robot acts based on imperfect perception.

\subsection{Simulation Environment and Asset Preparation}
All experiments are conducted in the PyBullet simulator. We utilize the object meshes from the YCB-Video dataset and nine distinct, widely-used robotic end-effector models provided by the \texttt{burg-toolkit}\footnote{\href{https://mrudorfer.github.io/burg-toolkit/}{https://mrudorfer.github.io/burg-toolkit/}} \cite{rudorfer2022}: the Franka Hand, Robotiq 2F-85, Robotiq 2F-140, WSG 32, WSG 50, EZGripper, Sawyer Hand, Kinova 3F, and Robotiq 3F.

The physics simulation is systematically controlled. For each trial, two instances of a single YCB-V object are spawned: a ground-truth (GT) version rendered in green with physics enabled, and an estimated (EST) version in red that serves only as a visual reference as seen in fig \ref{fig:pipeline}. To isolate interactions, the friction coefficient for the physics-enabled GT object is set to 0.5. The simulation runs at a high frequency of 240 Hz with 100 solver iterations to accurately model contact-rich scenarios. Critically, no ground plane is used; instead, objects are held in a floating environment by their pose constraints. Gravity is strategically managed: it is initially disabled during gripper positioning and closing to prevent premature object movement and is only enabled at $-9.81 \, m/s^2$ after the gripper has fully closed on the object to test for a stable lift.

\subsection{Evaluation Protocol}

Our evaluation protocol is designed to answer a central question: how does 3D model accuracy impact robotic grasping performance? In a practical scenario, a robot uses a reconstructed 3D model for two key tasks: to estimate the 6D pose of an object and to generate a set of viable grasp poses. The robot then executes one of these grasps on the real-world object. 

\subsubsection{Experimental Conditions}
To systematically analyze the different sources of error in this pipeline, we evaluate performance under three distinct conditions. In all cases, the grasp is executed on the ground-truth (GT) object model in the simulator, representing the physical reality.

\begin{itemize}
    \item \textbf{GT Model for Grasps \& Pose (Ideal Baseline):} 
    This condition establishes the best-case performance with "perfect" information. We use the ground-truth CAD model to both generate the library of canonical grasps and as the reference model for the 6D pose estimator. 

    \item \textbf{GT Model for Grasps \& Reconstructed Model for Pose (Isolating Pose Error):}
    This setup isolates the impact of a reconstructed model's geometry on \textit{pose estimation accuracy}. The robot plans grasps using a "perfect" internal GT model but uses the imperfect reconstructed mesh to find the object in the scene.

    \item \textbf{Reconstructed Model for Grasps \& Pose (End-to-End Realistic Scenario):}
    This represents the most realistic case. The robot uses the same imperfect, reconstructed 3D model for both generating grasp candidates and as the reference for pose estimation. This measures the compounded effect of errors from both stages.
\end{itemize}

\subsubsection{Trial Procedure}
The general procedure for each trial is as follows:
\begin{enumerate}
    \item For an object instance in a scene, retrieve its estimated pose $T_{c2o}^{est}$ and the pre-computed library of successful canonical grasps ($T_{o2g}$).
    \item For each canonical grasp, compute the target gripper pose using the estimated object pose via Eq.~\ref{eq:est_grasp}.
    \item Execute the grasp in the simulator on the object placed at its ground-truth pose.
    \item Record the outcome (success or failure) to calculate our functional metrics.
\end{enumerate}

\subsection{Evaluation Metrics}
\subsubsection{Grasp Generation Success Rate (\texorpdfstring{$S_{gen}$}{Sgen})}
\label{sec:metric_sgen}
This metric is used to evaluate the suitability of a 3D model for the grasp pose sampling stage. It measures the percentage of viable grasp candidates a model yields from a fixed set of randomly sampled grasp poses. 

Let $N_{total}$ be the total number of grasp poses sampled for an object. Let $N_{succ}^{\text{Model}}$ be the number of those poses that are successful when simulated on a specific \texttt{Model} (e.g., the ground-truth CAD model or a particular reconstructed mesh).

The \textbf{Grasp Generation Success Rate} for that model is:
$$
S_{gen}^{\text{Model}} = \frac{N_{succ}^{\text{Model}}}{N_{total}} \times 100\%
$$
 This metric directly quantifies how well a given 3D model's geometry supports the task of finding usable grasps.

\subsubsection{Estimated Success Rate (\texorpdfstring{$S_{est}$}{Sest})}
\label{sec:metric_sest}
This is the primary metric for evaluating grasping performance. It measures the probability that a grasp, which is known to be successful with a perfect object pose, will also succeed when using the pose provided by an estimation algorithm.

Let $G_{gt}$ be the set of all grasp poses that are successful when using the ground-truth object pose ($T_{c2o}^{gt}$). Let $N_{gt} = |G_{gt}|$ be the total number of such successful grasps. When these $N_{gt}$ grasps are executed using the estimated object pose ($T_{c2o}^{est}$), let $N_{succ}$ be the number of grasps that still succeed. The \textbf{Estimated Success Rate} is then defined as:
$$
S_{est} = \frac{N_{succ}}{N_{gt}} \times 100\%
$$

\subsubsection{Physics-Based Outcome Breakdown}
\label{sec:metric_breakdown}
To provide a detailed diagnosis of failures, grasp attempts in the physics simulation are categorized into the following outcomes:

\begin{itemize}
    \item \textbf{Successful Grasp:} The gripper successfully approaches, establishes a stable hold, and lifts the object against gravity without dropping it. This is the desired outcome.

    \item \textbf{Slipped:} A failure mode where the gripper makes initial contact but the hold is not stable, causing the object to slip from the grasp during the lift.

    \item \textbf{No Contact:} A failure mode where the gripper's fingers close completely without touching the object, typically caused by a large translation error in the estimated pose.

    \item \textbf{Collision:} A failure that occurs when the gripper's body collides with the object during approach, preventing a valid grasp from being attempted.
\end{itemize}

%%%%%%%%%%%%%%%%%%%%%%%%%%%%%%%%%%%%%%%%%%%

\section{Experiments}\label{sec:experiments}

\paragraph{Reconstruction Methods} 
To comprehensively evaluate the impact of 3D model fidelity, we utilize meshes generated by a diverse set of state-of-the-art techniques, sourced from the benchmark by Burde et al.~\cite{burde2024benchmark}. This selection spans multiple paradigms, including neural radiance field (NeRF) methods (Instant NGP \cite{instantngp}, NeRFacto \cite{Nerfstudio}, Neuralangelo \cite{Neuralangelo}), implicit surface models (UniSurf \cite{unisurf}, MonoSDF \cite{monosdf}, BakedSDF \cite{bakedsdf}, VolSDF \cite{volsdf}), and commercial photogrammetry software (RealityCapture \cite{capturereality}). Using this broad range of models allows us to analyze how different types of geometric inaccuracies and artifacts affect downstream grasping performance.

\subsection{Dataset and Pose Estimators}
We ground our benchmark in a real-world, challenging dataset and evaluate state-of-the-art object pose estimation methods.
\paragraph{Dataset} We use the YCB-Video (YCB-V) \cite{YCBV} dataset from the BOP challenge \cite{hodan2018bop}. It is known for significant clutter, occlusion, and lighting variations, and contains 21 objects with diverse geometries, sizes, and symmetries.

\paragraph{Pose Estimation Methods} We evaluate two leading pose estimators:
\begin{itemize}
    \item \textbf{MegaPose \cite{megapose2022}:} A powerful render-and-compare pipeline that achieves strong zero-shot performance through coarse-to-fine refinement.
    \item \textbf{FoundationPose \cite{foundationpose2023}:} A unified framework that integrates detection, tracking, and pose estimation, designed for robust performance on novel objects.
\end{itemize}
For every object instance in the YCB-V test set, we use the poses provided by these methods as the input $T_{c2o}^{est}$ to our evaluation pipeline.

\section{Results}\label{sec:results}

% Our large-scale simulation experiments reveal a clear hierarchy of how perception errors impact manipulation.
% We first isolate the effect of pose estimation error, then the error from imperfect 3D models used for sampling grasp poses, and finally analyze their combined, compounding effect.
Our experiments are designed to untangle how reconstructed geometry and the pose estimation errors impact manipulation. We structure our analysis in the following progression: first, we establish a baseline for gripper performance to understand the physical constraints of the task. Second, we isolate the effect of 6D pose estimation error on grasping success. Third, we isolate the effect of 3D model inaccuracies on grasp planning. Finally, we analyze the compounded effect of both pose and model errors to draw conclusions about their relative importance in a realistic end-to-end scenario.

\subsection{Baseline Gripper Suitability}

\begin{figure*}[!t]
\centering
\includegraphics[width=\textwidth]{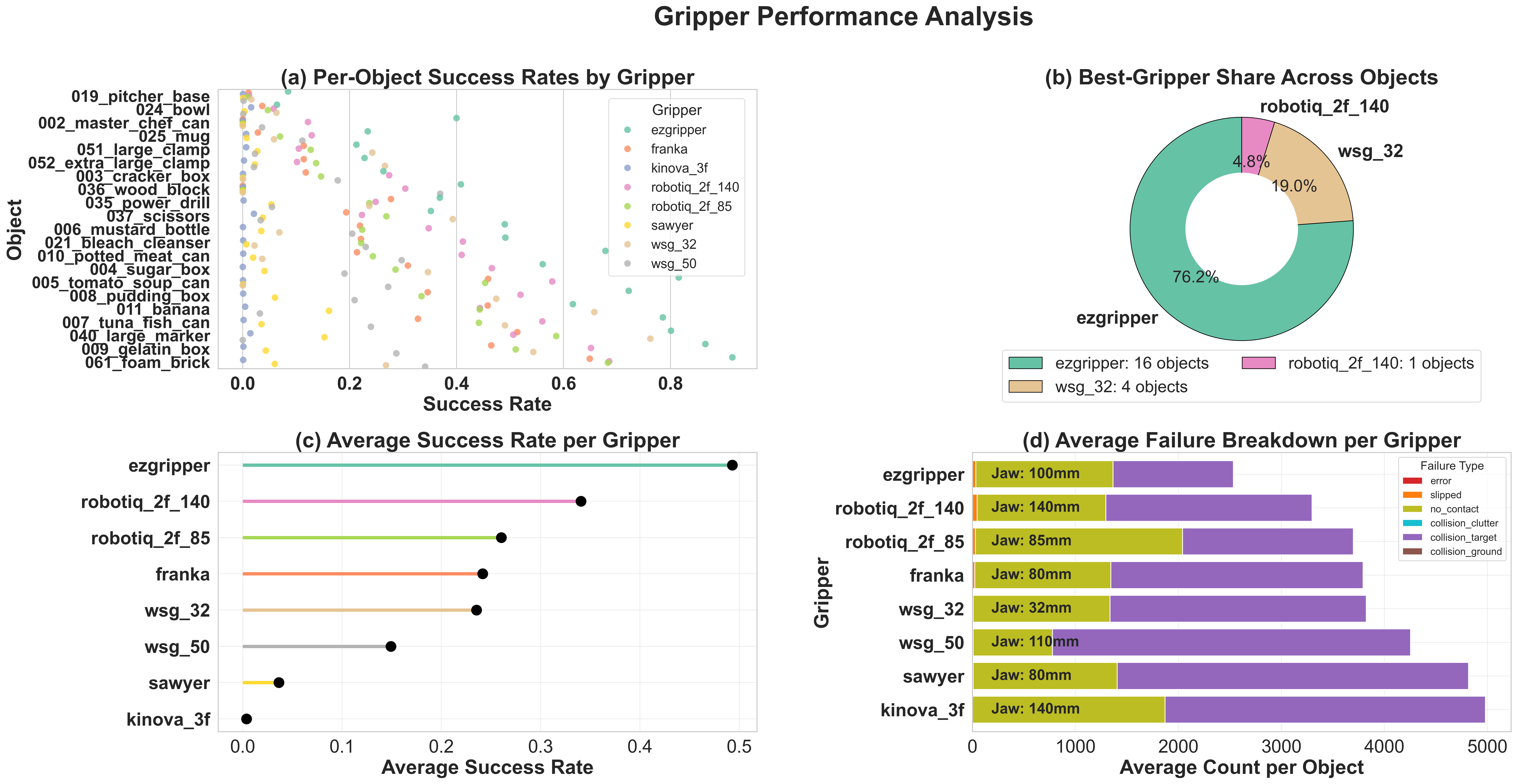}
% \caption{This figure presents a comprehensive analysis of gripper performance across 21 objects and 8 grippers in our grasp 
% benchmark. (a) The top-left strip plot shows per-object success rates for each gripper, with objects sorted by average success rate. (b) The top-right pie chart illustrates the distribution of best-performing grippers across all objects. (c) The bottom-left plot displays average success rates per gripper, sorted from lowest to highest. (d) The bottom-right stacked horizontal bar chart breaks down average failure types per gripper (collision with ground/target/clutter, no contact, slipping, and errors), with gripper jaw widths annotated in millimeters}
\caption{Baseline gripper performance analysis, visualizing the Grasp Generation Success Rate ($S_{gen}$) (Sec.~\ref{sec:metric_sgen}) across various grippers and objects under ideal conditions. (a) Per-object $S_{gen}$ for each gripper. (b) Distribution of the best-performing gripper for each object. (c) Average $S_{gen}$ per gripper. (d) A Physics-Based Outcome Breakdown (Sec.~\ref{sec:metric_breakdown}) of grasp failures for each gripper with gripper jaw widths annotated in millimeters.}
\label{fig:gripper_analysis}
\end{figure*}

To understand the impact of gripper design, we first established a baseline of each end-effector's innate capabilities. To do this, we performed a large-scale analysis for each of the 21 object-meshes and 9 gripper models. We sampled 5,000 diverse antipodal grasps for each pair on the object at its canonical identity pose ($T_{o}^{gt} = I$) and executed them in simulation. Figure~\ref{fig:gripper_analysis} presents this analysis, showing the performance of different grippers on the ground-truth object models at the canonical pose.
\begin{itemize}
    \item \textbf{Subfigure (a)} plots the Grasp Generation Success Rate ($S_{gen}$) (Sec.~\ref{sec:metric_sgen}) for each gripper on a per-object basis. It clearly shows that no single gripper is optimal for all objects; performance depends highly on the object's geometry.
    
    \item \textbf{Subfigure (b)} supports this by showing that the "best" gripper is distributed across several models, with the Robotiq 2F-85 and WSG 50 being the most frequent top performers.
    
    \item \textbf{Subfigure (c)} aggregates performance, showing the average $S_{gen}$ for each gripper across all objects, providing a general sense of which designs are most versatile.
    
    \item \textbf{Subfigure (d)} breaks down the failure modes (detailed in Sec.~\ref{sec:metric_breakdown}) for each gripper. We observe that grippers with smaller jaw widths often fail due to collisions, while wider grippers are more prone to slipping.
\end{itemize}
The primary conclusion from this baseline analysis is critical for our subsequent experiments: since gripper choice heavily influences success, relying on a single end-effector could bias our findings. Therefore, to ensure our conclusions about perception systems are generalizable, all the following experiments aggregate results across the entire library of grippers.

% Before assessing the impact of perception errors, it is essential to understand the inherent grasping capabilities of different end-effectors. Our baseline analysis, summarized in \textbf{Fig.~\ref{fig:gripper_analysis}}, provides this foundation by evaluating nine distinct grippers on all objects under ideal perception conditions. The results reveal significant variability in performance across different gripper morphologies and object geometries. For instance, the analysis highlights that no single gripper is optimal for all objects and quantifies the maximum achievable success rate for each object–gripper pair. Fig. \ref{fig:gripper_analysis} shows that a larger jaw size does not necessarily result in better grasping performance, while the three-finger gripper, which is more complex to implement due to finger actuation, shows the worst performance result. This baseline is crucial, as the following results on perception errors are aggregated across this diverse set of capabilities, ensuring that our conclusions about pose and reconstruction fidelity are not biased by the choice of a single, potentially unsuitable, end-effector.

\subsection{Object Pose estimation Error to Grasping Success}

\begin{figure*}[!t]
\centering
\includegraphics[width=\textwidth]{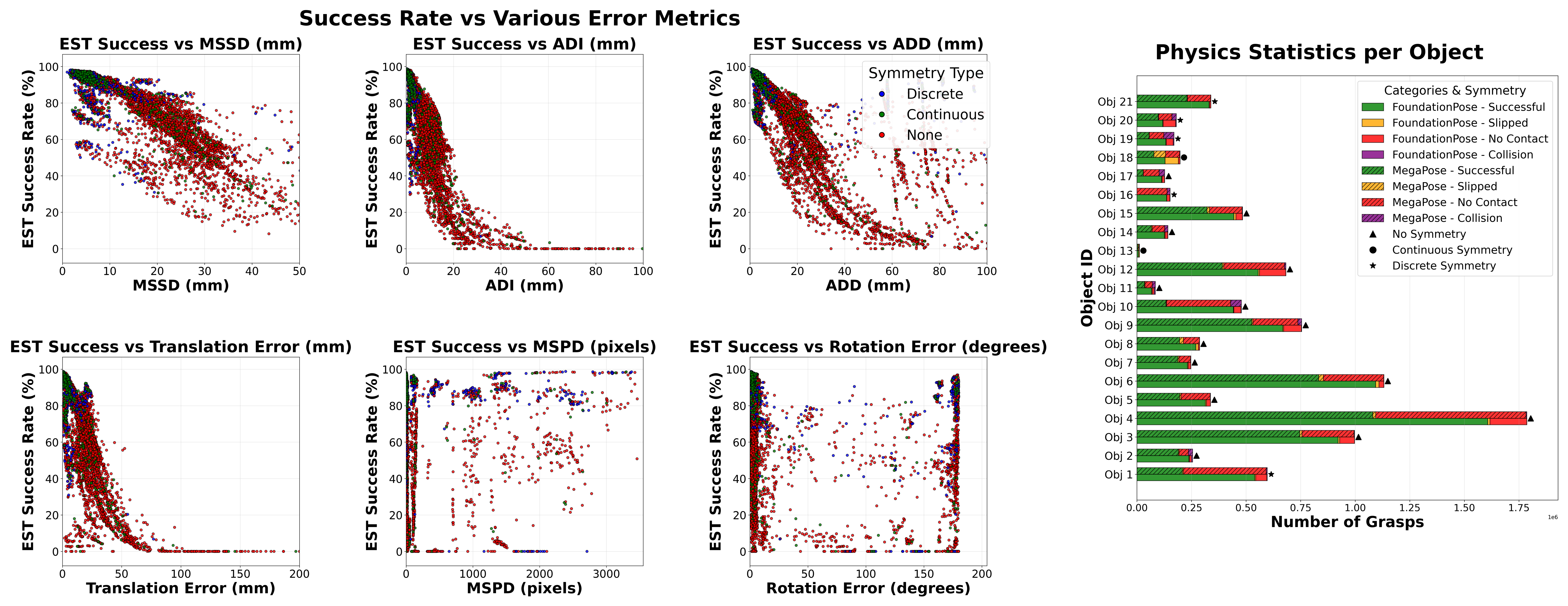}
% \caption{\textbf{Comprehensive Pose Estimation and Grasping Benchmark Analysis.} \textbf{Left Panel:} Scatter plots showing relation between estimated success rates and six pose estimation error metrics for both FoundationPose and MegaPose across 8,250 trials averaged over 18,882,842 simulation. Data points are color-coded by object symmetry. The plots reveal how pose estimation accuracy directly impacts grasping success. \textbf{Right Panel:} Physics-based grasping performance per object, comparing FoundationPose (solid colors) and MegaPose (hatched patterns). Each bar shows the proportional breakdown of simulation outcomes: Successful (green), Slipped (orange), No Contact (red), and Collision (purple).}
\caption{\textbf{Analysis of Grasping Performance vs. Pose Estimation Error.} \textbf{Left Panel:} Scatter plots showing the relation between various pose error metrics and the Estimated Success Rate ($S_{est}$), averaged over both FoundationPose and MegaPose across 8,250 trials and 18,882,842 simulations (Sec.\ref{sec:metric_sest}). \textbf{Right Panel:} A detailed Physics-Based Outcome Breakdown (Sec.\ref{sec:metric_breakdown}) of grasp attempts per object. The green portion of each bar represents the final $S_{est}$, while other colors show the proportions of different failure modes.}
\label{fig:comprehensive_analysis}
\end{figure*}

A key question is how well standard geometric metrics for pose estimation predict grasping task success. Figure~\ref{fig:comprehensive_analysis} addresses this by correlating pose errors with our functional metric, the Estimated Success Rate ($S_{est}$) (Sec.~\ref{sec:metric_sest}), under the ideal GT$\to$GT condition.

The \textbf{right panel} provides a high-level summary of grasping performance per object. It shows that a more accurate pose estimator leads to better grasping success. The green bars, representing the final $S_{est}$, are consistently taller for FoundationPose (89.9\% average success) than for MegaPose (59.4\%). The failure breakdown reveals why: MegaPose results in a much higher proportion of 'No Contact' and 'Slipped' failures, indicating its pose errors are often large enough to cause the gripper to miss the object or fail to secure a stable hold.

The \textbf{left panel} provides deeper insight by plotting $S_{est}$ against six different geometric error metrics. The distinct downward trend in the plots for 3D spatial metrics (MSSD, ADD, ADI, and translation error) demonstrates a strong correlation: as the 3D error increases, the probability of a successful grasp decreases. Conversely, the plots for 2D projection error (MSPD) and pure rotation error are much flatter, showing they are poor predictors of grasping success. This finding is significant as it validates that many standard 2D-based metrics do not capture the information most critical for physical interaction, highlighting the necessity of a benchmark like ours.

\subsection{The Impact of 3D Model Fidelity on Grasp sampling}

\begin{figure*}[!t]
\centering
\includegraphics[width=\textwidth]{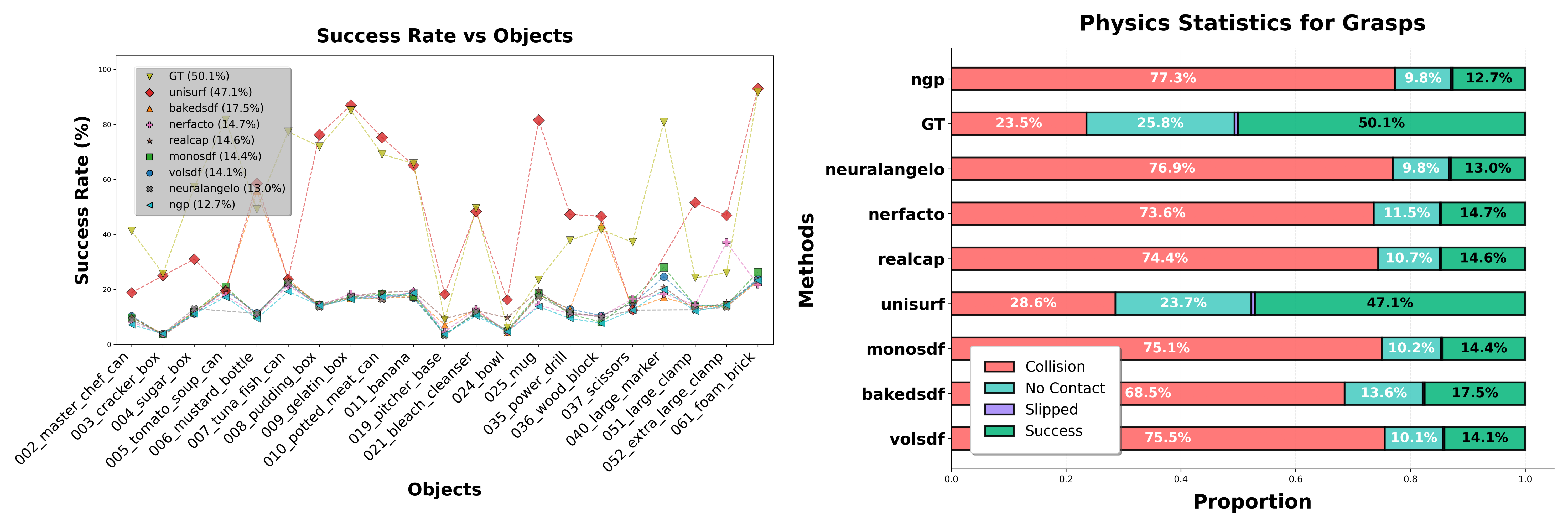}
% \caption{\textbf{Comprehensive Analysis of Grasp Pose Generation Performance and Physics-Based Failure Statistics.} \textbf{Left panel:} A performance overview showing success rate trends across different 3D reconstruction methods and YCB objects. \textbf{Right panel:} A breakdown of physics-based simulation outcomes. The stacked bars reveal that the quality of the geometric mesh critically influences grasping success. Lower-quality meshes significantly increase collision failures, demonstrating that mesh fidelity directly correlates with the ability to generate viable, collision-free grasp poses.}
\caption{\textbf{Impact of 3D Model Fidelity on Grasp Candidates.} \textbf{Left panel:} The Grasp Generation Success Rate ($S_{gen}$) (Sec.~\ref{sec:metric_sgen}) for various reconstruction methods. \textbf{Right panel:} A Physics-Based Outcome Breakdown (Sec.~\ref{sec:metric_breakdown}) for grasps planned on these meshes. Note the significant increase in 'Collision' failures for lower-quality models.}
\label{fig:failure_mode_analysis}
\end{figure*}

Next, we isolate the effect of geometric inaccuracies on finding the grasp candidates. To remove pose estimation from the equation, we generate grasps on various 3D models at their canonical identity pose. This directly measures how the quality of a 3D model affects the number of viable grasp candidates that can be generated.

The results are presented in Figure~\ref{fig:failure_mode_analysis}. The \textbf{left panel}, which plots the Grasp Generation Success Rate ($S_{gen}$) (Sec.~\ref{sec:metric_sgen}), shows a clear performance degradation for most reconstructed models compared to the ground-truth GT model. This means that geometric flaws and artifacts significantly reduce the number of valid grasp poses found by the sampler.

The \textbf{right panel} reveals the primary reason for this degradation. For lower-quality models like Instant-NGP, the dominant failure mode (Sec.~\ref{sec:metric_breakdown}) is 'Collision'. This indicates that the grasp sampler, operating on a flawed mesh, generates grasp poses that physically collide with the object geometry, making them unusable.

Interestingly, the Unisurf model yields an $S_{gen}$ comparable to the ground-truth model. A visual inspection of the Unisurf meshes reveals that they are often smoother and have fewer high-frequency artifacts than other reconstructions. This suggests that complex or noisy geometry can be more detrimental to the number of successful grasp candidates than a smoother, slightly less detailed representation.

\subsection{Compounded Errors: Pose and Geometry Inaccuracies Combined}
\begin{figure*}[!t]
\centering
\includegraphics[width=\textwidth]{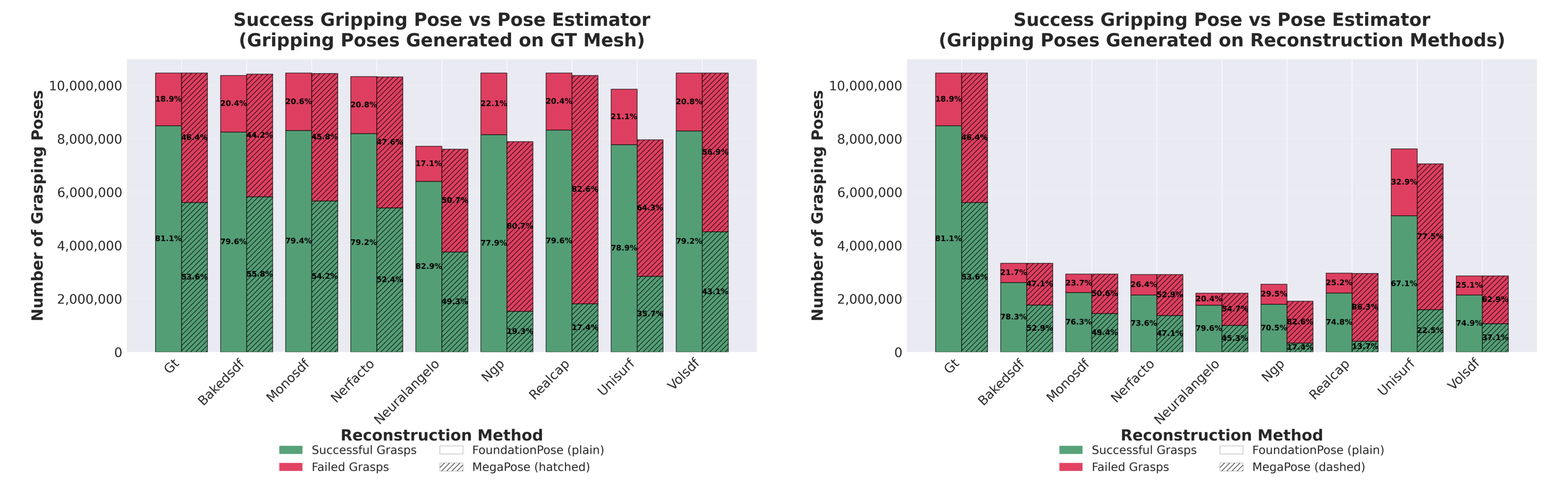}
% \caption{\textbf{Comparative analysis of grasp success rates under combined uncertainty.} \textbf{Left:} Grasp poses generated on ground truth (GT) meshes and evaluated on estimated object poses using the reconstructed mesh for object representation, showing baseline performance degradation due solely to pose errors. \textbf{Right:} Grasp poses generated directly on various reconstructed meshes and evaluated on estimated poses output from the pose estimator using the same mesh for object representation, demonstrating the combined impact of both reconstruction and pose estimation errors. Stacked bars show successful (green) and failed (red) grasps for FoundationPose (solid) and MegaPose (hatched). This comparison quantifies how mesh quality and pose accuracy jointly affect robotic grasping performance.}
\caption{\textbf{Comparative Analysis of Grasping Success under Compounded Errors.} This figure compares the final grasping success, measured by the Estimated Success Rate ($S_{est}$) (Sec.~\ref{sec:metric_sest}), when combining different sources of geometric and pose uncertainty. \textbf{Left:} Performance under the 'GT $\to$ Reconstructed mesh' condition. \textbf{Right:} Performance under the 'Reconstructed mesh $\to$ GT' condition.}
\label{fig:combined_error_analysis}
\end{figure*}

Finally, we analyze the end-to-end realistic scenario, where an imperfect reconstructed model is used to find grasp candidates and estimate pose. Figure~\ref{fig:combined_error_analysis} compares this compounded error condition (right panel) against a baseline where only pose estimation uses the reconstructed model (left panel).

The results lead to our main conclusion. As established in the previous section, using a reconstructed mesh drastically reduces the number of available grasp candidates. However, Figure~\ref{fig:combined_error_analysis} shows that as long as a sufficient number of candidates remain, the final grasping success, measured by $S_{est}$ (Sec.~\ref{sec:metric_sest}), is not significantly impacted for a high-quality pose estimator like FoundationPose. The green success bars in the right panel are slightly shorter than those in the left. This demonstrates that while 3D model fidelity is critical for generating a rich set of grasp options, the accuracy of the 6D pose estimate is the primary factor determining the ultimate success of the manipulation task.

% The most realistic scenario involves a robot using an imperfect, reconstructed model for both sampling grasp candidates and pose estimation reference. This introduces compounded errors from both model infidelity and pose estimation inaccuracy.

% Fig.~\ref{fig:combined_error_analysis} presents the results of this final experiment. The left panel shows the baseline performance under the \textbf{GT $\to$ GT} condition, where failures are due only to pose estimation error. The right panel shows the performance under the more challenging \textbf{Recononstructed mesh $\to$ GT} condition. The results are striking: a significant drop in the number of successful grasping poses, but surprisingly, a slight drop in grasping success is observed across all reconstruction methods for both FoundationPose and MegaPose. This result shows that the mesh quality can degrade the number of successful grasping poses while sampling, but the better pose estimation performance outweighs the grasping success over reconstruction quality. This side-by-side comparison unequivocally demonstrates that while pose estimation accuracy is critical, the quality of the underlying 3D object model is an equally important, and compounding, factor in the success of robotic grasping.

%%%%%%%%%%%%%%%%%%%%%%%%%%%%%%%%%%%%%%%%%%%

\section{Conclusion}
In this work, we introduced a benchmark to bridge the gap between standard geometric evaluation of perception systems and their grasping performance in robotic manipulation. Our results provide a clearer understanding of how 3D reconstruction and object pose errors affect grasping success.

Our quantitative analysis reveals a strong correlation between 3D object pose error and the probability of a successful grasp. While minor pose errors are often tolerated, we observed that success drops sharply once the error exceeds a certain threshold. This confirms that while standard 3D metrics are informative, a direct evaluation is necessary to understand the practical limits of a perception system. Furthermore, our analysis of different experimental conditions shows how using an imperfect reconstructed model as a reference for pose estimation degrades performance, even when grasp candidate sampling is performed on a perfect internal model.

Beyond the accuracy of the 6D object pose, we revealed that the fidelity of the reconstructed 3D model critically impacts the sampling of grasp candidates. Grasping performance deteriorates significantly when grasp candidates are generated on imperfect reconstructions, primarily because geometric artifacts in the mesh lead to increased planned grasps, resulting in collisions with the actual object.

Finally, our end-to-end analysis clarifies the codependent relationship between model quality and pose accuracy. A high-quality mesh is the foundation for success, as it is required to generate a rich set of viable grasp candidates and enable an accurate 6D pose estimate. However, our results show that the final accuracy of the object's pose is the more direct determinant of grasping success. A state-of-the-art pose estimator can often compensate for moderate geometric inaccuracies in its reference model. Still, even a perfect pose cannot recover a grasp that was miscalculated on a severely flawed mesh. This highlights that while mesh quality is foundational, pose estimation accuracy is the more dependent metric for successful manipulation.

The primary limitation of our work is its dependence on simulation. Future work will focus on validating these findings on a physical robotic platform. Furthermore, this framework can be extended to manipulation primitives beyond grasping, such as investigating how the gripper pose uncertainty impacts high-precision placement and assembly tasks. Ultimately, this work argues for a shift in how we evaluate perception systems in robotics, supporting benchmarks that account for the entire perception-to-action pipeline.

\section*{ACKNOWLEDGMENT}
The authors acknowledge the use of the Gemini AI tool to refine and improve the text and grammar. We also acknowledge the assistance of Claude and Grok in refining the Python code used to generate the plots.

% Add the bibliography
\bibliographystyle{IEEEtran}
\bibliography{refrences}

\newpage
\onecolumn % Optional: Use this if you want the appendix to span the full page width
% \twocolumn % Keep this if you want to maintain the two-column format
%%%%%%%%%%%%%%%%%%%%%%%%%%%%%%%%%%%%%%%%%%%

% \section*{Supplementary Material}

% \subsection{Implementation Details}
% To support the reproducibility of the benchmark presented in Section \ref{sec:experiments}, we provide the exact simulation parameters and asset specifications used in PyBullet.

%%%%%%%%%%%%%%%%%%%%%%%%%%%%%%%%%%%%%%%%%%%

\end{document}